\documentclass{article}

\usepackage{arxiv}

\usepackage[utf8]{inputenc} % allow utf-8 input
\usepackage[T1]{fontenc}    % use 8-bit T1 fonts
\usepackage{hyperref}       % hyperlinks
\usepackage{url}            % simple URL typesetting
\usepackage{booktabs}       % professional-quality tables
\usepackage{amsfonts}       % blackboard math symbols
\usepackage{nicefrac}       % compact symbols for 1/2, etc.
\usepackage{microtype}      % microtypography
\usepackage{amsmath,amssymb,amsfonts}
\usepackage{cleveref}       % smart cross-referencing
\usepackage{ulem}
\usepackage{graphicx}
\usepackage{natbib}
\usepackage{xcolor} 
\usepackage{doi}
\usepackage{subcaption}
\usepackage{array}
\usepackage{multirow}

\title{Modulating Reservoir Dynamics via Reinforcement Learning for Efficient Robot Skill Synthesis}

\newif\ifuniqueAffiliation
\uniqueAffiliationtrue

\ifuniqueAffiliation % Standard variant of author block
\usepackage{authblk}

\newbox{\orcid}\sbox{\orcid}{\includegraphics[scale=0.06]{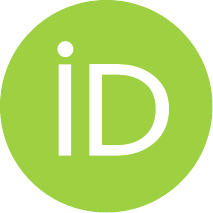}} 
\author[1]{%
	\href{https://orcid.org/0000-0000-0000-0000}{\usebox{\orcid}\hspace{1mm}Zahra Koulaeizadeh\thanks{\texttt{zahra.koulaeizadeh@ozu.edu.tr}}}%
}
\author[1,2]{%
	\href{https://orcid.org/0000-0002-3051-6038}{\usebox{\orcid}\hspace{1mm}Erhan Oztop\thanks{\texttt{erhan.oztop@otri.osaka-u.ac.jp}}}%
}
\affil[1]{Department of Artificial Intelligence, Ozyegin University, Istanbul, Turkey}
\affil[2]{OTRI/SISReC, Osaka University, Osaka, Japan}
\fi

\hypersetup{
pdftitle={Modulating Reservoir Dynamics via Reinforcement Learning for Efficient Robot Skill Synthesis},
pdfsubject={q-bio.NC, q-bio.QM},
pdfauthor={Zahra Koulaeizadeh,Erhan Oztop},
pdfkeywords={Learning from Demonstration, Reinforcement Learning, Echo State Networks, Robot Learning},
}

\begin{document}
\maketitle

\begin{abstract}
	A random recurrent neural network, called a reservoir, can be used to learn robot movements conditioned on context inputs that encode task goals. The Learning is achieved by mapping the random dynamics of the reservoir modulated by context to desired trajectories via linear regression. This makes the reservoir computing (RC) approach computationally efficient as no iterative gradient descent learning is needed. In this work, we propose a novel RC-based Learning from Demonstration (LfD) framework that not only learns to generate the demonstrated movements but also allows online modulation of the reservoir dynamics to generate movement trajectories that are not covered by the initial demonstration set. This is made possible by using a Reinforcement Learning (RL) module that learns a policy to output context as its actions based on the robot state. %Since the reservoir output weights are not changed, each RL action causes a variational change in the whole movement trajectory. 
Considering that the context dimension is typically low, learning with the RL module is very efficient. We show the validity of the proposed model with systematic experiments on a 2 degrees-of-freedom (DOF) simulated robot that is taught to reach targets, encoded as context,  with and without obstacle avoidance constraint. The initial data set includes a set of reaching demonstrations which are learned by the reservoir system. To enable reaching out-of-distribution targets, the RL module is engaged in learning a policy to generate dynamic contexts so that the generated trajectory achieves the desired goal without any learning in the reservoir system. Overall, the proposed model uses an initial learned motor primitive set to efficiently generate diverse motor behaviors guided by the designed reward function. Thus the model can be used as a flexible and effective LfD system where the action repertoire can be extended without new data collection. 
\end{abstract}

% keywords can be removed

\section{Introduction}
Learning-from-demonstration (LfD) has become an effective method to equip robots with movement skills in diverse tasks, from basic manipulation to complex navigation \cite{schaal1999imitation, argall2009survey, chernova2014robot}. By generalizing from human demonstrations, LfD enables robots to replicate intricate behaviors with minimal manual programming. This capability has made LfD an attractive approach in robotics, especially for applications requiring rapid learning and adaptation \cite{urain2024deepgenerativemodelsrobotics, ravichandar2020recent}. In behavior cloning (BC)\cite{behavior_cloning}, a common LfD method, the robot learns to map observed states to demonstrated actions, enabling it to achieve task goals based on previous examples.

However, Behavior Cloning (BC) methods are highly dependent on the training data distribution, making them susceptible to covariate shift problem\cite{behavior_cloning}, where discrepancies between the training and deployment data lead to compounding errors over time\cite{accumulation_err}, thereby limiting their ability to generalize to out-of-distribution (OOD) tasks and novel targets.
To cope with this situation, often BC methods need to be retrained with new data or augmented with additional learning modules, as in ACNMP\cite{acnmp}. However, these approaches create an additional computational burden. Here, we also follow an augmentation approach but keep computational efficiency as a critical design factor. To be concrete, we propose a novel model, the Dynamic Adaptive Reservoir Computing model (DARC), which combines the efficiency of Reservoir Computing (RC) \cite{Reservoir_Computing} with the adaptability of Reinforcement Learning (RL) \cite{712192}. We maintain  learning efficiency by having RL to operate on low dimensional action spaces.

Reservoir Computing, particularly in the form of Echo State Networks (ESNs) \cite{echo}, has gained attention for temporal sequence learning due to its computational efficiency. Unlike traditional recurrent neural networks that require iterative weight updates, ESNs use a randomly initialized recurrent layer (the reservoir) with fixed dynamics, while only the output weights are trained via linear regression. This structure makes ESNs highly efficient for real-time control tasks. However, ESNs alone struggle to extrapolate beyond their training set, limiting their applicability in environments requiring novel, adaptive behaviors. 

In our approach, the reservoir network is fixed after an initial demonstration phase, where it learns a set of motor primitives. Instead of updating the reservoir weights, we use an RL module that learns a policy to dynamically generate context inputs for the reservoir. This allows the system to adapt to OOD tasks without additional demonstrations or reservoir modifications, extending the robot's action repertoire.
%and enabling online adaptation to new tasks or environmental constraints. 
By generating dynamic context inputs through the RL module, our approach offers a lightweight solution to the  generalization problem in LfD, where obtaining new demonstration data may be costly or impractical. This RL-based training is independent of the initial demonstration phase, distinguishing our approach from other methods that require retraining with initial data. %, such as ACNMP
\cite{acnmp}. In sum, the contributions of this work are:

\begin{itemize}
    \item Improved LfD flexibility due to the modulation of reservoir activity through context inputs, allowing the model to generate novel behaviors by leveraging the stage one learning without additional data collections.
    \item Computational Efficiency due to (1) fast stage 
 one training thanks to reservoir learning, and (2) learning in low dimensional action space with RL in the second stage.   
    \item  Improved scalability due to the decoupling of task complexity from context dimensionality, allowing generalization to more complex tasks with limited computational cost increase in RL.
\end{itemize}

\section{Related Work}

% \subsection{Learning from Demonstrations}

Reservoir Computing (RC) \cite{Reservoir_Computing} is a learning framework that utilizes a fixed random recurrent neural network(RNN)\cite{medsker2001recurrent}, called the reservoir, to transform a given input or input sequence into a high-dimensional dynamical system, which then can be mapped to a desired temporal sequence via linear regression. Since only the output layer is trained, RC is adopted as a lightweight learning system applicable to time series prediction and control tasks, where temporal dependencies are critical\cite{reservoircomputingroboticsreview}. %This makes it particularly useful for systems  
One of the most commonly used implementations of Reservoir Computing (RC) is the Echo State Network (ESN), introduced in \cite{echo}. The reservoir of an ESN is required to have the so called \emph{echo state property}, which ensures that the internal states respond transiently to inputs and gradually fade, providing temporal stability and short-term memory. This property has proven effective not only for time-series tasks, such as those in \cite{time_series_1, time_series_2, time_series_3}, but also for robotic applications.

The memory capacity and simplicity of ESNs have been effectively utilized in robotic control tasks, as demonstrated in \cite{esn_memory_enhanced, esn_robot_controller}, where ESNs serve as controllers, and in \cite{esn_autonomous}, where an ESN is used to generate motor trajectories. In this latter application, the ESN is combined with a perception module and a controller to achieve adaptive trajectory generation. A recent extension of ESNs, known as Context-based Echo State Networks (CESNs), has an   additional context input, that may encode target position, obstacle size, etc. so that trajectories are generated conditioned on the context.

\begin{figure*}[t] % t indicates the figure should be placed at the top
\centering
\includegraphics[width=0.8\textwidth]{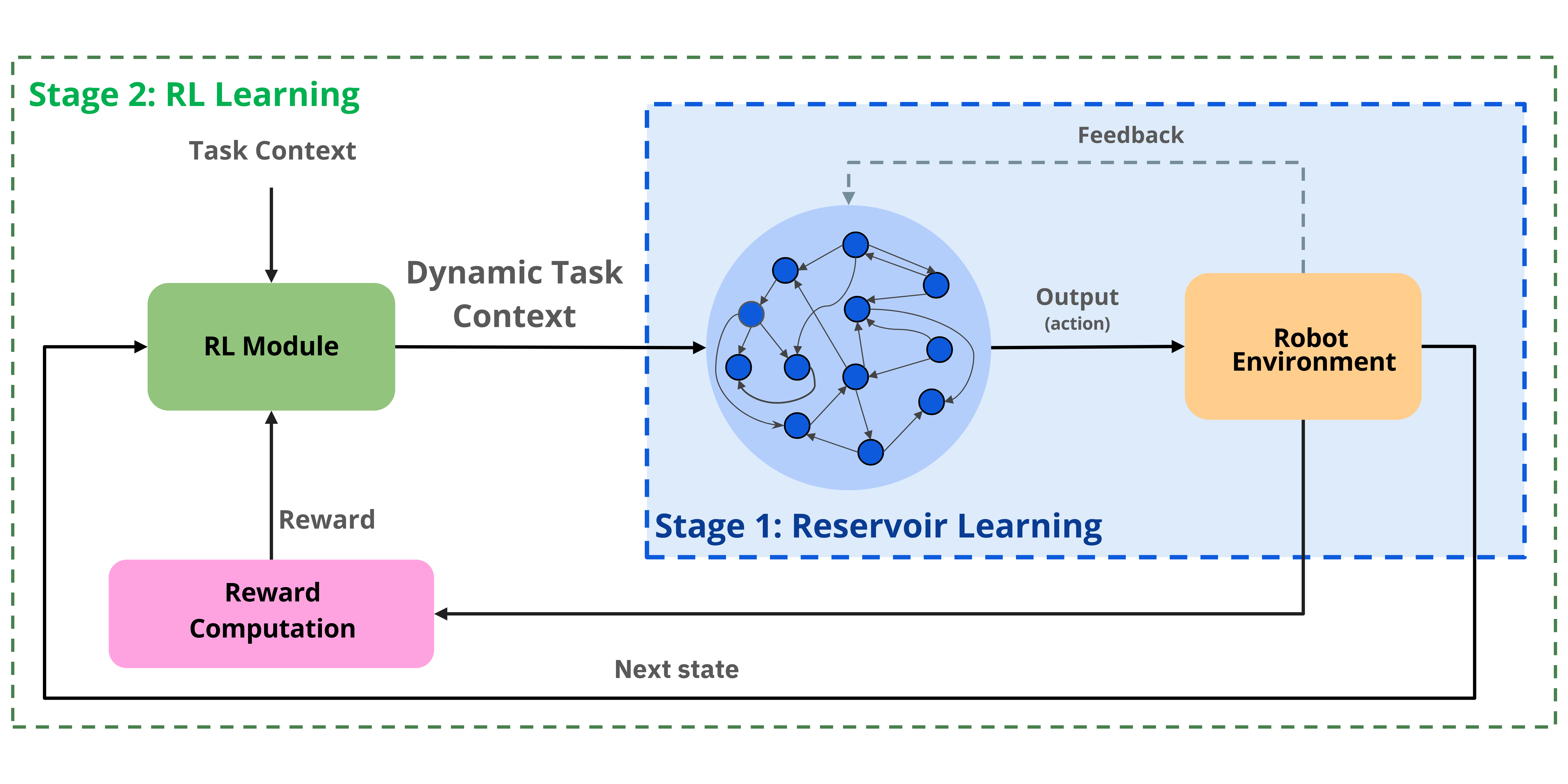} % Adjust width as needed
    \caption{Schematic of the DARC (Dynamic Adaptive Reservoir Computing) model. In the model, the Reservoir, as well as the Input, Context and Feedback weights are fixed and initialized to appropriate random values. In the first stage, the reservoir is trained by using an initial set of tasks (context, movement trajectory pairs) yielding an output weight matrix, which is fixed thereafter. For an out-of-distribution variant of the learned task or a completely novel task,  Stage 2 is engaged where the RL module builds a policy to make online context modulations so that the desired new task goal, captured by the  Reward function, can be achieved. After RL is completed, DARC model is able perform the initial task as well as the targeted new task with no additional reservoir training.
    }
    \label{fig:rl_cesn_architecture}
\end{figure*}

Recent works, such as Kawai et al. \cite{kawai2023reservoir} and Kawai et al. \cite{kawai2024oscillationsenhancetimeseriesprediction}, have attempted to address instability and generalization issues in reservoir computing. In the first study [20], the authors proposed the reBASICS (reservoir of basal dynamics) model to handle tasks with long-term intervals. reBASICS consists of multiple small reservoirs, where the reduced network size helps mitigate instability by minimizing the impact of chaotic dynamics. This approach effectively stabilizes the system, allowing it to reproduce learned time-series data with greater consistency. However, despite this improvement, the model show difficulties in generalizing to new time-series patterns, limiting its capacity to handle unfamiliar scenarios\cite{kawai2023reservoir}.  

In the second study \cite{kawai2024oscillationsenhancetimeseriesprediction}, the authors introduced random sinusoidal oscillations as input signals during task execution to address both instability and generalization issues. This method was effective for the specific tasks tested, as the oscillations stabilized the reservoir dynamics and improved generalization within controlled settings. However, these tasks were relatively simple and lacked direct relevance to robotics applications.
While the oscillatory input was effective in these controlled experiments, it is unclear if this approach would be sufficient for real-world robotic tasks, where the system may need task dependent online input adjustments rather than fixed oscillatory resorvior stimulation.

To extent the capabilities of RC-based  learning, researchers have combined it with reinforcement learning (RL). In cite{reinforcementlearningconvolutionalreservoir}, RL is used to adjust reservoir output weights for better decision-making in control tasks, while in \cite{deepqnetworkusingreservoir}, RL tunes the reservoir’s internal parameters to adapt to complex dynamics. However, both approaches have high computational costs and sample inefficiency, limiting their use in real-time robotic systems, as high number of parameters are tuned by RL. In contrast, our Dynamic Adaptive Reservoir Computing (DARC) model reduces the RL task action dimensionality, thereby offering a more efficient, lightweight solution with faster convergence.

The combination of Learning from Demonstration (LfD) and Reinforcement Learning (RL) approaches can be broadly divided into two categories. In the first category, RL serves as the primary learning framework, with demonstrations used to guide and accelerate the RL process by providing expert examples \cite{lfd_rl_1_1,lfd_rl_1_2,lfd_rl_1_3,lfd_rl_1_4}. For instance, Rajeswaran et al. \cite{lfd_rl_1_3} incorporate human demonstrations within a deep RL framework to improve learning efficiency for complex manipulation tasks. While this approach reduces sample complexity, it still encounters challenges in fine-tuning for varied real-world conditions and maintaining stability, particularly when scaling to more intricate tasks or adapting to new scenarios not seen during training.

The second category, which our work belongs, uses LfD as the primary learning mechanism, with RL employed to enhance and extend its capabilities. Our approach is similar to\cite{acnmp, kober2012reinforcement}.
In \cite{acnmp}, Adaptive Conditional Neural Movement Primitives (ACNMP) integrate LfD with RL by first learning a shared latent representation of movement trajectories via Conditional Neural Processes (CNPs). This representation is then used by RL to generate adapted control policies that optimize the robot’s actions for specific task goals in new environments. However, a drawback of ACNMP is the computational cost of retraining the CNMP model with each optimized trajectory, which can limit its use and scalability in resource-constrained settings.
In \cite{kober2012reinforcement}, reinforcement learning is applied to adapt motor primitives to new situations by adjusting a set of global parameters, yet the approach’s  focus on tuning a fixed set of meta-parameters for specific tasks could hinder exploration efficiency and generalization when applied to highly diverse, unseen tasks.

Compared to the mentioned works, our model Dynamic Adaptive Reservoir Computing (DARC) differs in that it combines the efficiency of reservoir computing with RL-generated low-dimensional dynamic context inputs to improve the LfD-learned skills as well as exploiting the existing behavior repertoire to acquire novel task without new data collection and experience replay of the initial LfD data.

\section{Methodology}
Our motivation is to use a high-dimensional dynamical system that is open to modulation via reinforcement learning in order to expand the trajectory generation capacity of the original system. For this, as the basis network, Context-based Echo State Network (CESN) model \citep{cesn} is used as the LfD model. We call the LfD learning by CENS as the stage-1 learning. Stage-2 learning involves RL, where stage-1 learning weights are kept fixed and RL module learns to modulate context inputs of CESN to achieve a desired novel task or novel task goal. Stage-2 learning is explained in subsection~\ref{sub:Methods_RL}.

%which is shown to be able learn a set of trajectories conditioned on a fixed set of context inputs . We first present it in the next subsection, then explain how it is used in our model together with RL.

\subsection{Context based Echo State Networks(CESN)}\label{AA}

A CESN consists of an input layer, a reservoir, and an output layer. The input layer may contain an input time series, feedback from the from the, possibly controlled, environment, and contextual information capturing global task goals.  The reservoir is a recurrent neural network with fixed weight which is connected to the input layer through fixed weights.  The output layer includes linear read-out layer that is trained to generate desired target trajectories for the given inputs and context conditions. 
We extend the CESN model by allowing the \emph{context} to be a dynamical variable that can depend on the reservoir and/or environment state, and use a slightly different formalization by having separate  input layer weights. We denote the reservoir state (unit activities) at time t with $x(t)$, the control input with $u(t)$, the feedback with $f(t)$, and the context with $c(t)$. Then the reservoir is updated at discrete time steps by using the following equations.
\begin{align}
\mathbf{x}_{net} &= \mathbf{W}_x \mathbf{x}(t-1) 
+ \mathbf{W}_{\text{c}} 
\mathbf{c}(t)
+ \mathbf{W}_{\text{f}} \mathbf{f}(t) 
+ \mathbf{W}_{\text{u}} \mathbf{u}(t) \label{eq:res_net_in} && \\
\mathbf{x}(t) &= (1 - \alpha) \mathbf{x}(t-1) + \alpha {\text{tanh}(\mathbf{\mathbf{x}_{net} })} \label{eq:res_update} &&
\end{align}
Where $\mathbf{W}_x \in \mathbb{R}^{N_r \times N_r} $, $\mathbf{W}_{\text{c}} \in \mathbb{R}^{N_r \times N_c }$, $\mathbf{W}_{\text{f}} \in \mathbb{R}^{N_r \times N_y}$, and $\mathbf{W}_{\text{in}} \in \mathbb{R}^{N_r \times N_u}$ are the reservoir, context, feedback, and the input weight matrices respectively. $\alpha \in (0, 1]$ is a smoothing parameter softening the change in reservoir state at each time step. $N_r$ is the reservoir size determined by the designer. $N_u$, $N_c$, and $N_y$ are determined by the learning task and indicates the input command ($u$), context $c$, and the output ($y$) dimensions respectively. 

Given a desired output time series $\mathbf{y}(t)$ with corresponding input $\mathbf{u}(t)$ and context $\mathbf{c}(t)$, one can set up a linear regression problem to map reservoir states $\mathbf{x}(t)$ to output $\mathbf{y}(t)$, through a linear read-out weight matrix $\mathbf{W}_{\text{out}}$,  which can be approximately solved with ridge regression with $\lambda$ as the ridge parameter:
\begin{align}
    %\mathbf{W}_{\text{out}} = \mathbf{Y}\mathbf{X}^\dagger
    \mathbf{W}_{\text{out}} = \mathbf{Y} \mathbf{X}^T (\mathbf{X} \mathbf{X}^T + \lambda \mathbf{I}_m)^{-1}
\end{align}
 where the rows of $\mathbf{Y}$ are made up of $\mathbf{y}^T(t)$ and the rows of $\mathbf{X}$  are constructed from the reservoir state as $[1,\mathbf{x}^T(t)]$, for $t=1,2,...,T$, where $T$ is the end time. Once having computed the $N_y \times (N_r + 1)$ output matrix $\mathbf{W}_{\text{out}}$, one can do predictions given a reservoir state $\mathbf{x}(t)$ with
\begin{align}
\mathbf{y}^T(t) &= \mathbf{W}_{\text{out}} 
\begin{bmatrix}
1, \mathbf{x}^T(t)
\end{bmatrix} \label{eq:readout_eq}
\end{align}
After this first stage, the computed the $\mathbf{W}_{\text{out}}$ matrix together with other fixed random weights are saved to be used in RL-based training in stage-1I.

\subsection{Dynamic context with Reinforcement Learning}
\label{sub:Methods_RL}
 In the second stage of DARC training (see Figure~\ref{fig:rl_cesn_architecture}), the purpose is to train the RL module to generate modulated contexts for the extrapolated task. The problem is formalized as a Markov Decision Process (MDP) represented as a tuple \(<S, A, T, R>\), where \(S\) is the state space, \(A\) is the action space, defined here as a subset of \(\mathbb{R}^{N_c}\), with \(N_c\) being the context dimension. The state $s$ $\in S$ can be taken as the state of the controlled system and/or reservoir state $\mathbf{x}$ (Equation~\ref{eq:res_update}).  $T$ represents the transition function or the transition probability distribution that describes how states evolve. In our case, it is directly determined by the reservoir dynamics given in Equation~\ref{eq:res_update}. Finally, $R$ defines the reward function, which in our setting is designed to reward the generation of a desired new behavior, which may be an improvement in accuracy over the extrapolated version of the original task or even the generation of a completely novel task behavoir. As usual, the RL objective is to obtain a policy \(\pi_{\theta} : S \times A \rightarrow \mathbb{R}^{N_c}\) that learns how to modulate the dynamics of the reservoir network by outputting the context input as a function of state to maximize the expected episodic return or the cumulative discounted reward, $J$:
\[
J(\pi_{\theta}) = \mathbb{E}_{\pi_{\theta}} \left[ \sum_{t=0}^{T} \gamma^t R(s_t, a_t) \right],
\]
\noindent
where \(\gamma \in [0,1]\) is the discount factor determining the importance of future rewards. The reward function $R$ may be defined as the sum of a running reward \( R_\text{run} \) and a terminal reward \( R_{\text{ter}} \), where \( R_{\text{t}} \) provides incremental rewards at each time step to encourage task progression, and \( R_{\text{f}} \) is the final reward given at the completion of the task.

Thus, the action generated by the policy $\pi_{\theta}$, being the context \(\hat{c}_t\), is injected into the reservoir dynamics through the $\mathbf{W}_{\text{c}} 
\mathbf{c}(t)$ component of reservoir dynamics Equation~\ref{eq:res_net_in}.

\section{Test Platform and Tasks Addressed}

\subsection{Environment}

The experiments were carried out in the Reacher environment of Gymnasium OpenAI \citep{towers2024gymnasium}, which is widely used as a benchmark for reinforcement learning tasks involving robotic arms. The environment features a 2 degrees-of-freedom (DOF) robotic arm attached to a fixed base (see Figure~\ref{fig:env}b). The objective is to generate joint torques to make the arm reach the desired target points in the 2D plane (see Figure~\ref{fig:env}a).

\subsection{Evaluation Tasks}

Since DARC is a hybrid model that combines Learning from Demonstration (LfD) and Reinforcement Learning (RL), we evaluate its performance against two component baselines to understand the contributions of each core element: the CESN\citep{cesn} model as an LfD-only approach, and a standalone Proximal Policy Optimization (PPO)\citep{ppo} agent as an RL-only approach. The CESN model used here is identical to the one integrated into our DARC model, allowing us to evaluate the effectiveness of dynamic context generation provided by the RL module. Similarly, we trained the standalone PPO agent under the same conditions as our RL module within DARC, including the action generation frequency, reward structure, and number of training episodes. By analyzing DARC alongside these baselines, we can assess the contributions of each component and demonstrate the added value of combining LfD with RL for enhanced generalization and adaptability. For evaluations we designed three experiments as detailed next. 
%To do so, we designed three experiments consisting of task variations of increasing complexity. These tasks enable us to systematically evaluate each model’s performance in terms of accuracy, adaptability, and efficiency under varying conditions.

\subsubsection{\textbf{Reaching task}}
In the first task, the objective is to control the arm to reach a series of predefined target points scattered throughout the workspace. These target points differ from those used to train the reservoir, challenging DARC to generate appropriate trajectories for unseen locations. Performance is evaluated based on how closely the arm approaches each target and the efficiency of task completion.

\subsubsection{\textbf{Reaching while avoiding obstacle task}}
In the second task, we increased the difficulty of the task by introducing an obstacle among the target points as shown in Figure~\ref{fig:env}.a. In this task, the arm has to follow the trajectory without colliding with the obstacle. The goal is to assess the model’s capacity for obstacle-aware trajectory generation 
% \erhcom{RL, LfD distinction?}
, requiring the agent to modulate its path to avoid collisions while still reaching the target. 
% \erhcom{Same comment as above}

\subsubsection{\textbf{Novel Task of circular path tracking}}
% \erhcom{do not call this Extreme Extrapolation: "Extrapolation to Novel Task" or simply "Novel Task" is better. So I guess the 2 tasks above are also extrapolation tasks, is that correct? If so make it clear in the text. LfD learns reaching, works good with interpolation but fails with extrapolation, then we engage stage-1I and see how it works etc..}
In this third task, the goal is to evaluate DARC’s ability to perform transfer learning. Specifically, we use the reservoir model trained in the second task (reaching with an obstacle) but now train the RL module to track a moving target along a circular path. This setup tests whether DARC can adapt to a new, dynamic task by leveraging the knowledge gained from the previous task with static targets. The DARC model’s performance is evaluated based on its ability to generate accurate trajectories that follow the path while avoiding the obstacle.

\subsection{Data Collection}
\label{sec:data_collection}
\begin{figure}[t]
    \centering
    \includegraphics[width=0.6\textwidth]{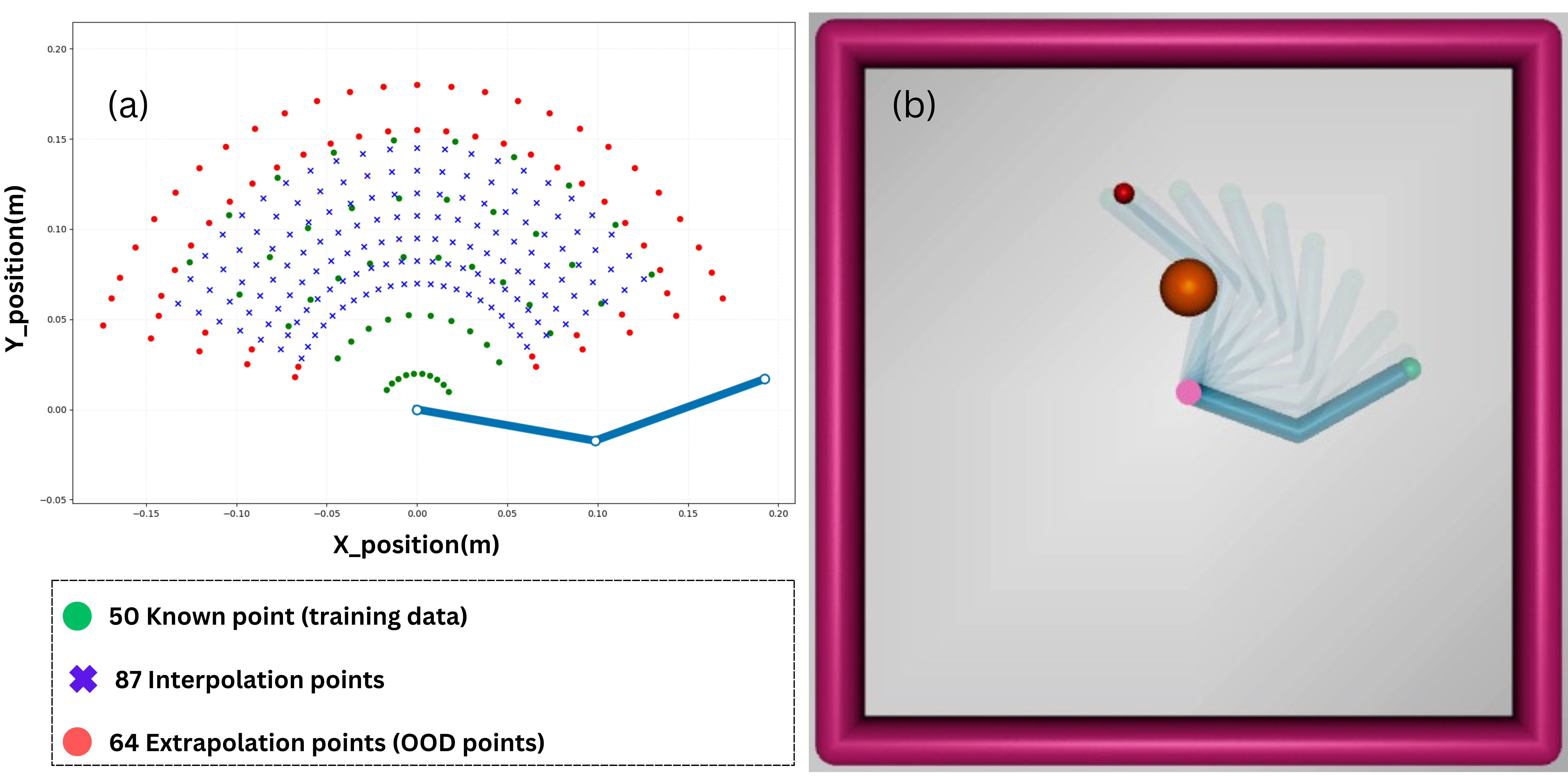}
    \caption{Experimental setup: (a) Target distribution with known,
interpolated, and extrapolated points. (b) Simulation of the 2-DOF robotic arm in the Reacher environment, showing trajectory towards the target while avoiding the obstacle.}
    \label{fig:env}
\end{figure}

In the first stage of training the DARC model, we utilized a feedback controller to collect the dataset 
\(\mathcal{D} = \{ \mathcal{D}_t^{(e)} \mid t = 1, \dots, T, \, e = 1, \dots, E \}\), where each episode \(e\) contains a sequence of data points \(\mathcal{D}_t^{(e)} = (c_t^{(e)}, \tau_t^{(e)}, f_t^{(e)})\) over \(T\) timesteps. Here, \(\mathbf{c}_t^{(e)}\) is the context data, specifically the target position \(P_{\text{target}}(t)\) with Cartesian coordinates \([p_{t_x}, p_{t_y}]\); \(\mathbf{\tau}_t^{(e)} = [P_{\text{end effector}}(t), u_1(t), u_2(t)]\) represents the trajectory data, including the end effector position \(P_{\text{end effector}}(t) = [p_{e_x}, p_{e_y}]\) and joint torques \(u_1(t)\) and \(u_2(t)\); and \(f_t^{(e)} = [q_1(t), q_2(t), \dot{q}_1(t), \dot{q}_2(t)]\) provides the feedback data, comprising joint angles and velocities.
This dataset \(\mathcal{D}\) provides the structured inputs required for training the reservoir model. Training data points are shown in Figure~\ref{fig:env} by green dots.

\section{Implementation Details}

\subsection{Stage-1: Learning an initial set of movement with LfD}

We adopt the core framework of the CESN model to perform initial training on a set of demonstrated task trajectories coupled with their task parameters, i.e., contexts. To ensure the \textbf{echo state property}, the reservoir weights \( W_x \) are scaled by an appropriate factor\citep{lukovsevivcius2012practical}. Additionally, the feedback weights \( W_f \) and context weights \( W_c \) are scaled to fine-tune the reservoir's performance\citep{lukovsevivcius2012practical}.

In this implementation, the input to the reservoir at each timestep \( t \) consists of the task-specific context \(\mathbf{c}_t = [x_{\text{target}}, y_{\text{target}}]\). For the first task (reaching task), which is relatively simple, we only use the context \(\mathbf{c}_t\) without feedback. However, for the more complex tasks (reaching with obstacle avoidance and circular trajectory generation), we incorporate the feedback signal \({f}_t = [q_1(t), q_2(t), \dot{q}_1(t), \dot{q}_2(t)]\), representing joint angles and velocities, to improve performance. For further increasing the robustness of the system, the feedback signal \(\mathbf{f}_t\) is augmented with noise(see Appendix\ref{sec:A} for details).
% so that  $\hat{\mathbf{f}}_t = \mathbf{f}_t + \boldsymbol{\epsilon}$ where \(\boldsymbol{\epsilon}\) is a noise vector with components drawn from a uniform distribution  with range and amplitude tuned according to the variability in the demonstration data set. 
% \erhcom{what are the actual values? Report them somewhere, e.g. Appendix etc.} 
%
%\begin{equation}
%\epsilon \sim \mathcal{U}([b_l, b_h])
%\label{eq:uniform_noise}
%\end{equation}
%
%\noindent with \( b_l \) and \( b_h \) are selected based on the %natural variation in the dataset values, ensuring the noise range %reflects the inherent variability observed in the task demonstrations.

% \erhcom{(Connect here with one or two sentence to the collected data and the formulation used in Eq3, Eq4)} 

The context and feedback data collected in Section \ref{sec:data_collection} are utilized to generate the reservoir states using Equations~\ref{eq:res_net_in},\ref{eq:res_update}, and subsequently, the trajectory data from the dataset is employed to compute the output weights as described in Equation~\ref{eq:readout_eq}.
Once the output weights matrix $\mathbf{W}_\text{out}$ is computed by Equation~\ref{eq:readout_eq}, trajectories can be generated at each time step $t$ as \(\boldsymbol{\tau}_t = [p_x(t), p_y(t), u_1(t), u_2(t)]\) through matrix multiplication with the learned output weight ({Equation}~\ref{eq:readout_eq}). Thus the system outputs the torques $u_1(t), u_2(t)$ to drive the robot arm as well as its prediction for the endeffector position, $(p_x(t),p_y(t))$. So it can be used for torque control or PD-control of the robot. In the current implementation only torque outputs are used to drive the robot.
%The output weights \( W_{\text{out}} \) were obtained via ridge regression on the training data, minimizing the error between predicted and target trajectories.

\subsection{Stage-2: Dynamic context generation with RL}

In the second stage, the RL module is trained to solve the MDP task $(S, A, R, T)$, in which the state $\mathbf{s}_t \in S$ is  defined as $\mathbf{s}_t = [\cos(\mathbf{q}), \sin(\mathbf{q}), \mathbf{q}, \mathbf{\dot{q}}, \mathbf{p_t}, \mathbf{p_e}]$ where 
$\mathbf{q}$ and $\mathbf{\dot{q}}$ represent the angles and angular velocities of the joints of the robot respectively, and $\mathbf{p_t}$ represents the position of the target and $\mathbf{p_e}$ robot end effector position. 
% \erhcom{You were going to use $\theta$ or $\mathbf{q}$  for joints. Please use consistent notation for clarity}
% \erhcom{Question: you define the p as relative distance to target. When context is fixed target=context. What happens when context changes? Do you keep a copy of the real target position?}

%the cosine and sine of the two joint angles \(\cos(\mathbf{q}_{\text{pos}})\) and \(\sin(\mathbf{q}_{\text{pos}})\), the target position \(\mathbf{q}_{\text{pos}}\), the angular velocities of the joints \(\mathbf{q}_{\text{vel}}\), and the relative position vector between the target and the reacher’s current position \(\mathbf{x}_{\text{pos}}\). Thus we have %Formally, we denote the state vector as:
%\[
%\mathbf{s}_t = [\cos(\mathbf{q}_{\text{pos}}), %\sin(\mathbf{q}_{\text{pos}}), \mathbf{q}_{\text{pos}}, %\mathbf{q}_{\text{vel}}, \mathbf{x}_{\text{pos}}]
%\]

The action space \(A\) consists of continuous actions \(\mathbf{a}_t \in [-1, 1] \subset \mathbb{R}^{N_c}\), matching the dimensionality of the context input to the CESN reservoir (\(N_c\)). %In the DARC model, the RL module generates the actions every k step depending on the difficulty of the task. 
For this study, Proximal Policy Optimization (PPO)\citep{ppo} is chosen as the RL algorithm due to its effectiveness in continuous state and continuous action space problems. Note that technically PPO learns a stochastic policy but we treat it as a deterministic policy by always selecting the mean of the outputted Gaussian distribution.
In the DARC model, the RL module can be set to generate actions at every \(k\) steps, where \(k\) is a meta-parameter chosen according to task and computational requirements. A smaller $k$ indicates more frequent context changes, with minimum $k=1$ indicating that the context is updated at each reservoir state update.
%selected based on the complexity of the task. For simpler tasks, \(k\) may be larger (e.g., \(k = 25 
%,50\)), while for more challenging tasks, it is smaller (e.g., \(k = 5, 10\)), allowing more frequent updates to enhance control precision.

We employ a 4-layer architecture to implement the actor and critic networks of the PPO agent. Both the actor and critic networks use an experience replay buffer to store experience tuples \((\mathbf{s}_t, \mathbf{a}_t, r_t, \mathbf{s}_{t+1})\). At each update, a batch of 2000 transitions \(\{(\mathbf{s}_i, \mathbf{a}_i, r_i, \mathbf{s}_{i+1})\}_{i=1}^N\) stored in the replay buffer, is used to update the policy and value networks over 100 epochs using the Adam optimizer, with the learning rate set to 1e-3 for the actor network and 1e-4 for the critic network. 
% \erhcom{(what is the batch size?, what is the experience buffer size? Please report these here or in a Table in an Appendix. Table/Appendix is good if these are different for different Experiments)} 

% \erhcom{by using Adam optimizer with learning rate set to 1e-4 (ENTER THE CORRECT TEXT)}. 

% \erhcom{Is this suggested by a paper? "This process allows the PPO agent to learn from diverse experiences, stabilizing the training by reducing correlations between consecutive updates and improving sample efficiency." If yes, citep and say you follow their suggestion. Otherwise, you cannot say it stabilized and reduces correlations, since you have not shown in your data. If you want ot write this, make it as a motivational sentence, not a statement.}

\subsection{Reward Function} 
% \erhcom{Overall: please use plain writing (clean math and clean text), as I indicated before for methods. Avoid unnecessary bullets, : separators, paragraphs etc.}
To accelerate policy learning, we used reward shaping in the experiments reported in this study. A running reward function $R_{\text{run}}(t)$  is designed to guide the learning towards the desired solution, while a terminal reward $R_{\text{term}}(t)$ is used to indicate the level of completion of the task. A composite running reward function is used:
%The reward shaping used in the experiments consists of a combined reward, including a running reward \( R_t \) at each timestep and a terminal reward \( r_f \) at the end of the task.
%
%1. \textbf{Running Reward}: The running reward at each timestep \( t \) is calculated as:
%
\begin{equation}
   R_{\text{run}}(t) = \alpha r_d(t) + \beta r_s(t) +  r_o   \label{eq:reward}
\end{equation}
 \noindent where $r_d(t)$ is the Euclidean distance to the specified task goal in the current step, $r_s(t)$ is a bonus reward for staying close to the target defined by
 \[
r_s(t) = 
\begin{cases} 
+1, & \text{if } \| {P}(t) - {P}_{\text{desired}}(t) \| < 5 \, \text{mm} \\ 
0, & \text{otherwise} 
\end{cases}
\]
and $r_o(t) = -10$ when the arm is in collision with the obstacle otherwise it is zero.
% \erhcom{You gave the equation of Euclidean distance that is known to everyone, but you did not give how you compute $r_s$ and $r_o$. Are they one time rewards? What happens with collision? etc..}
% where:
%    - \( r_d \): Distance to Target—the Euclidean distance between the end effector and the target at timestep \( t \):
%      \begin{equation}
%               r_d = \| \mathbf{x}_{\text{end effector}}(t) - \mathbf{x}_{\text{target}} \|
%               \label{eq: dits_to_trgt}
%      \end{equation}

%    - \( r_s \): Stay Reward—a positive reward if the end effector stays close to the target, encouraging stability near the target point after reaching it.
%    - \( r_o \): Obstacle Penalty—a negative reward assigned if the end effector collides with an obstacle, applied only in the second and third tasks.

% \erhcom{(Please convert Terminal Reward part to plain text as I wrote for running reward. Do not uses "-" etc. Follow my writing style to be clear and compact.)}
The terminal reward \( r_f \) differs between the first two tasks and the third task. In the first and second tasks, if the task is successfully completed in the final step (i.e., the end effector reaches within 1 cm of the target), the terminal reward is
defined as $r_f = {c} - \gamma  L$ where ${c}$ is a positive constant and $L$ is the path length computed by $L = \sum_{t=1}^{T} \| {P}(t) - {P}(t-1) \|$ where ${P}$ is the end effector positions, and $T$ is the end time step. This term encourages the model to minimize the path-length while reaching the target.  
% \erhcom{(Decide whether to use bold p or italic p for end-effector position)}

Path-length minimization is not necessary for circular trajectory generation, as the goal is to follow a circular trajectory which the running reward should take care of. Instead, the terminal reward is defined as $ r_f = {c} - \eta  p_e
 $ where $\theta$  is a coefficient to weight the trajectory error, and \( p_e \) is the average Euclidean distance error between the followed trajectory and the desired circular trajectory, computed as $p_e = \frac{1}{T} \sum_{t=1}^{T} \| {P}(t) - {P}_{\text{desired}}(t) \|$, where $p_\text{desired}$ indicates desired end effector position.  
 % \erhcom{(You were going change $\theta$ here to a different symbol!)} 
 % \erhcom{(And how do you choose c and $\theta$?)}
 % \erhcom{(Below, you used x as a vector but before it was a scalar, and it was also used in CESN. Use the above notation)}
\noindent
   The coefficients \(\alpha\), \(\beta\), \(\gamma\), \(\theta\) and {c} are tuned based on task requirements to balance each component’s influence on the agent’s behavior. In the experiments presented in this study they are chosen empirically as shown in table \ref{tab:task_parameters}. 
   % \erhcom{(Then give $c$ too!)}

\newcolumntype{C}[1]{>{\centering\arraybackslash}p{#1}}
\begin{table}[ht]
\centering
\setlength{\tabcolsep}{0.7pt} % Adjust column padding for compactness
\renewcommand{\arraystretch}{0.9} % Adjust row height for readability
\begin{tabular}{C{0.1\linewidth} C{0.2\linewidth} C{0.2\linewidth} C{0.2\linewidth}}
\toprule
\textbf{Parameter} & \textbf{Reaching} & \textbf{Reaching with Obstacle} & \textbf{Circle Tracking} \\
\midrule
$\alpha$ & -1 & -0.5 & -0.01 \\
$\beta$  & 2 & 1 & 0.01 \\
$\gamma$ & 100 & 100 & - \\
% $\lambda$& 100 & 100 & - \\
$\eta$ & - & - & 100 \\
${c}$ & 100 & 100 & 10 \\
\bottomrule
\end{tabular}
\caption{Reward parameters for each task}
\label{tab:task_parameters}
\end{table}
\section{Results}
In the experiments conducted in this study, the CESN model was trained on 50 demonstrated target points for the reaching task. For the reaching task with an obstacle, target points within the obstacle region were excluded. However, additional points were included on the left side of the obstacle, as reaching this area is more challenging. Consequently, a total of 80 points were used for the second and third tasks. The model's generalization was then evaluated on 64 extrapolated target points that were not included in the training set(see Figure~\ref{fig:env}a).
% \erhcom{give the correct number for all three tasks 50, 100, ?}, 
For DARC, the same CESN model was further extended with an RL module, trained to adapt to these 64 new targets. Lastly, we evaluate a standalone RL model trained and tested directly on the 64 extrapolated points. For reaching tasks, the model performances are assessed with \emph{Final Distance to Target} and \emph{Path Length} measures.  A successful reaching is expected to have a low endpoint error avoiding convoluted trajectories. See Table~\ref{tab:train_config} for setup details of each experiment.

\begin{table}[ht]
\centering
\renewcommand{\arraystretch}{1.3}
\resizebox{\textwidth}{!}{%
\begin{tabular}{>{\bfseries}lcccccccccc}
\toprule
\multirow{2}{*}{} & \multicolumn{3}{c}{Output period(steps)} & \multicolumn{3}{c}{Number of Training Episodes} & \multicolumn{3}{c}{Training Time (minutes)} & \multirow{2}{*}{Total timesteps} \\
\cmidrule(lr){2-4} \cmidrule(lr){5-7} \cmidrule(lr){8-10}
& PPO & DARC & CESN & PPO & DARC & CESN & PPO & DARC & CESN & \\
\midrule
Reaching Task & 25 & 25 & 1 & 200k & 50k & 50 & 80 & 25 & $<1$ & 50 \\
Reaching with Obstacle Avoidance Task & 5 & 5 & 1 & 200k & 50k & 80 & 100 & 28 & $<1$ & 100 \\
Novel Task of Circular Path Tracking & 5 & 5 & 1 & 30k & 30k & 80 & 40 & 40 & $<1$ & 300 \\
\bottomrule
\end{tabular}%
}
\caption{Training configuration }
\label{tab:train_config}
\end{table}

\subsection{\textbf{Reaching task}}
This section first establishes how CESN performs on the novel test targets that are outside its training range. Then, it evaluates how successful our DARC model is in leveraging CESN in reaching those points with additional RL. In this experiment the RL module of DARC  is allowed to change context input only twice (i.e., k=25, where full trajectory length is 50). To assess the baseline difficulty of the task, we also train a PPO model to learn a policy to drive the robot endeffector to these test targets. Each model is trained twice with different random seeds. As each run includes 64 points, the models are evaluated on a total of 128 reaching episodes. 
\begin{figure}[t]
    \centering
    \includegraphics[width=0.6\textwidth]{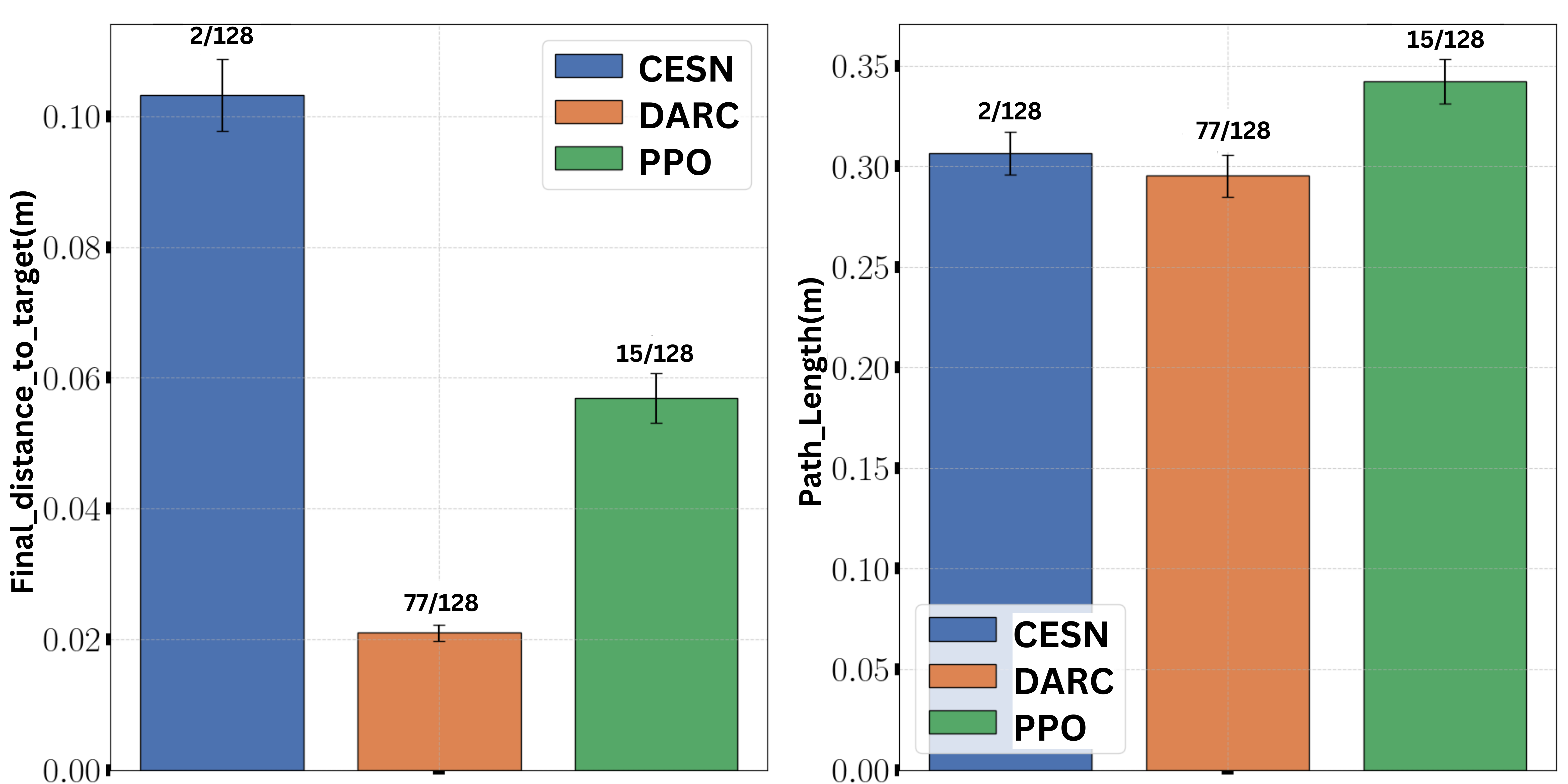}
    \caption{Reaching task test results over two training sessions for each model. (Left) Final distance to target (mean ± SEM), where DARC outperforms CESN and PPO with 77/128 successful reaches, followed by PPO with 15/128, and CESN with only 2/128. (Right) Path length to target shows that DARC maintains relatively shorter paths compared to CESN and PPO} % reflecting more efficient trajectory planning.}
    \label{fig:t-test_easy}
\end{figure}

\begin{figure}[ht]
    \centering
    \includegraphics[width=0.85\textwidth]{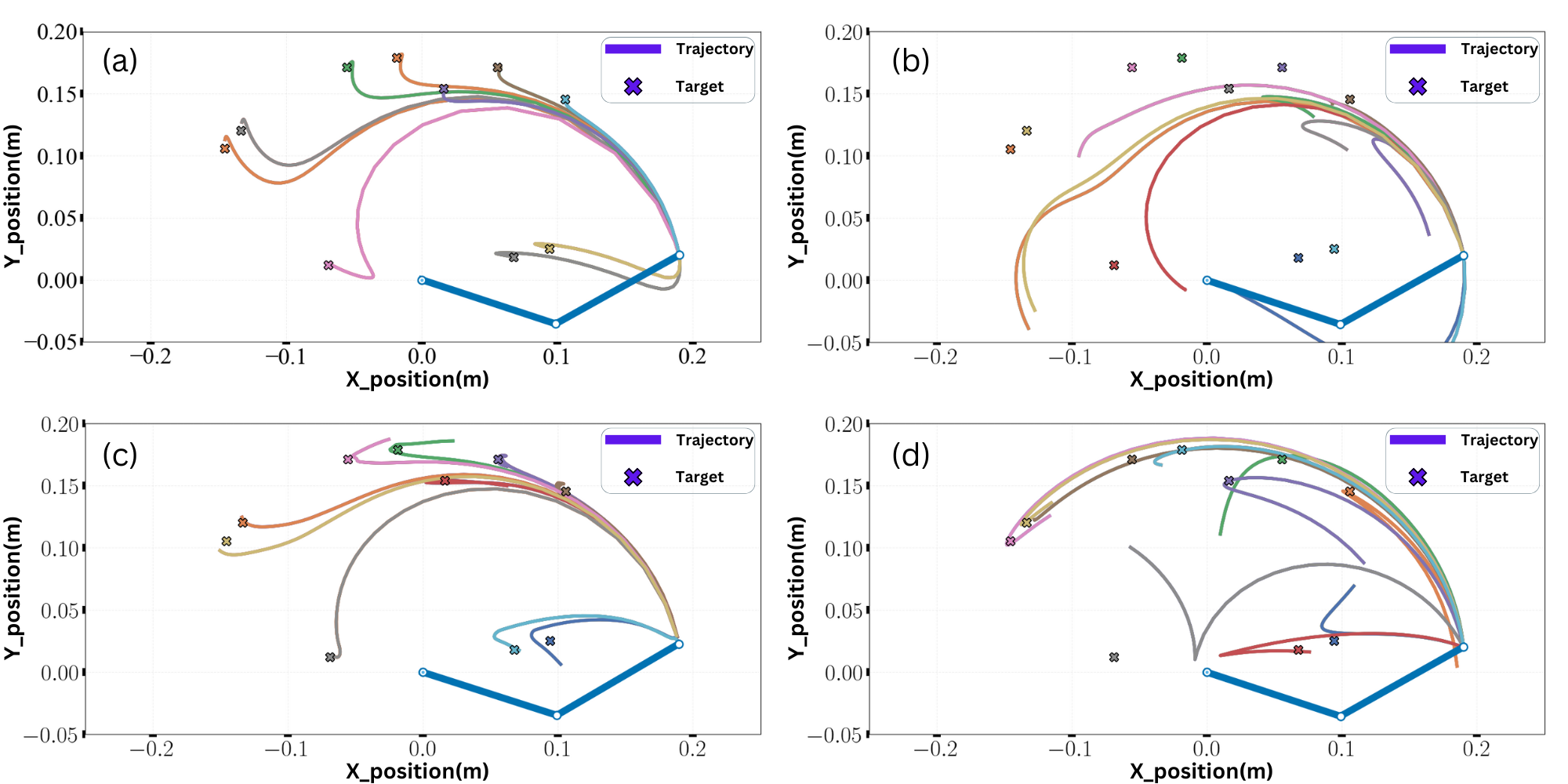}
    \caption{Comparison of 10 test trajectories for the Reaching task. (a) Ground truth from PD controller. (b) CESN model. (c) DARC model. (d) PPO model.}
    \label{fig:easy_trajs}
\end{figure}

As shown in Figure~\ref{fig:t-test_easy}, the DARC model outperforms both CESN and standalone PPO on both metrics. 
For the distance-to-target metric, DARC demonstrates significantly improved accuracy, achieving a success rate of 77 out of 128 episodes in which the robot can reach the target within 1 cm proximity. This performance is notably higher than CESN and PPO, which only reached the target in 2 and 15 episodes, respectively. Note that in this experiment we only allowed DARC to modify the context only twice during the whole execution. Even, this gave an impressive gain in performance. In terms of the path-length metric, DARC achieves a similar trajectory efficiency to CESN, as CESN provides the base behavior for DARC. Yet, dynamic context generation by RL, gives a slight advantage to DARC over CESN in terms of path-length. 

%\erhcom{You have to explain how the path-length calculated and averaged, for failed trials. Probably PPO would have much worse path length if it were allowed to run 50 steps}

To give an intuitive feel of how the robot behaves under the control of these three models, the trajectories generated by each model for a set of targets are given in Figure~\ref{fig:easy_trajs}. The CESN model (Figure~\ref{fig:easy_trajs}b) seems to make a correct start but cannot quite make it to the target points. Especially, it has difficulty for the points closer to the robot base. In contrast, the DARC model (Figure~\ref{fig:easy_trajs}c) reaches target points smoothly and accurately.
%by generating dynamic context input only twice during the episode. 
The standalone PPO model, shown in Figure~\ref{fig:easy_trajs}d, performs poorly, as expected since it is allowed to generate actions at every 25 steps.
% \erhcom{This is an overkill for PPO, it has no chance! (you ask it to give two or tree torque commands to reach the targets). Are the PPO result in Figure 3 also obtained with the same setting?? I guess not, because it reached 15/128 times. Let's make these two cases consistent. Also here you must tell about how long PPO and DARC was allowed to run. If you used the same procedure you can define them earlier and not repeat here.You can also make table to report all experiment settings}. %The reduced action frequency prevents PPO from making the precise adjustments required to reach the targets, leading to incomplete and irregular trajectories. This result emphasizes the computational efficiency of our proposed DARC model. By dynamically generating context only twice per episode, DARC achieves effective target-reaching behavior while addressing the long training time typically associated with reinforcement learning algorithms.

\subsection{\textbf{Reaching with obstacle avoidance task}}

In this section, we assess the performance of CESN, DARC, and PPO models on the second task,
which requires reaching target points in 100 steps while avoiding obstacles.
% \erhcom{write a similar sentence for the first experiment as well. The goal is to reach in 50 steps}. 
%This experiment examines each model's adaptability to constraints, specifically its ability to adjust trajectories to avoid collisions and accurately reach targets. 
In this experiment, each model is tested over 4 training runs on 64 points, resulting in a total of 256 test episodes.

\begin{figure}[ht]
    \centering
    \includegraphics[width=0.6\textwidth]{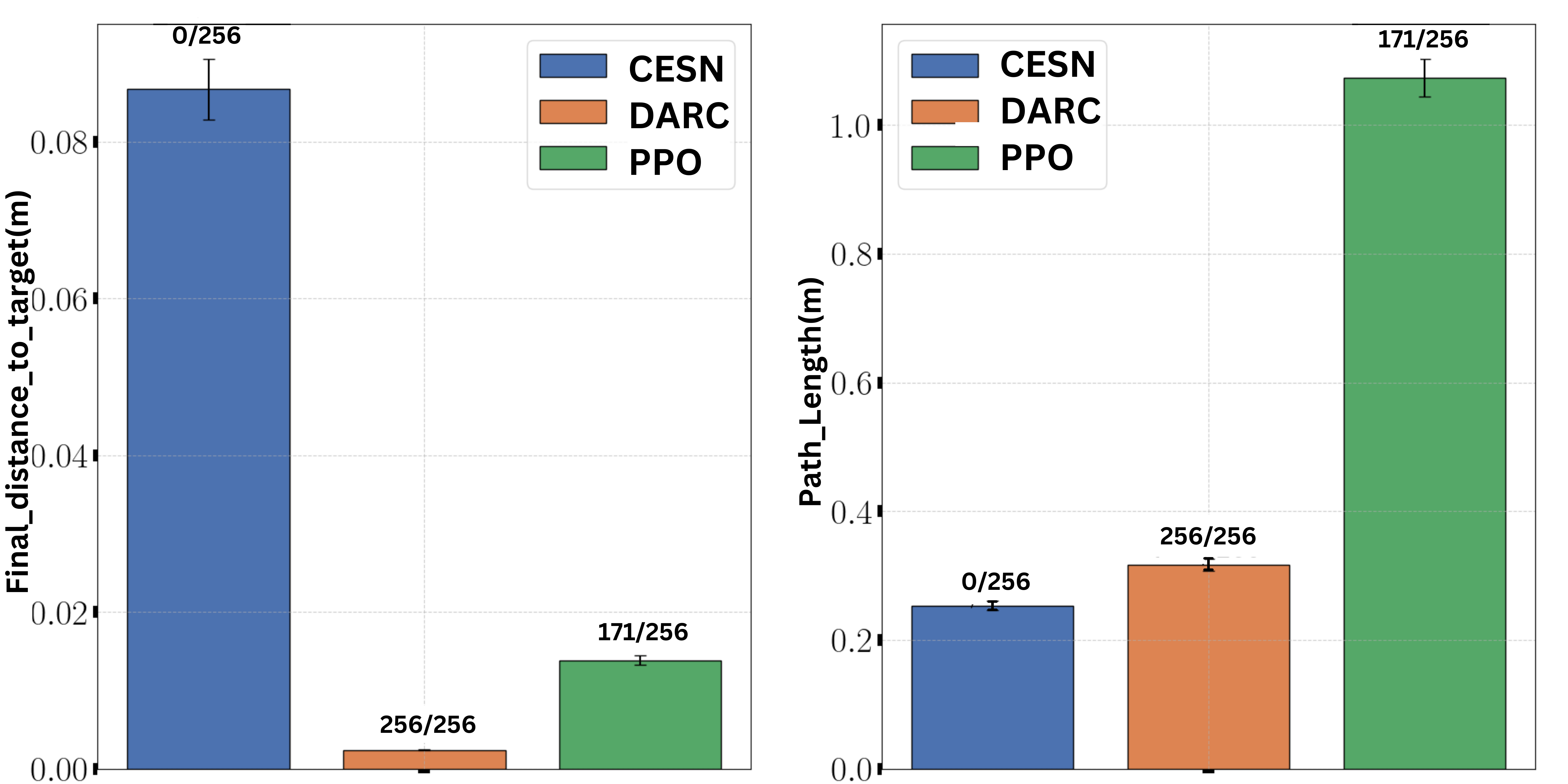}
    \caption{Test results over four training sessions for each model for Reaching with Obstacle Avoidance task . (Left) Final distance to target (mean ± SEM) shows that DARC achieves the most accurate positioning with all targets reached (256/256), followed by PPO, which successfully reaches 171 out of 256 targets. CESN has the largest final distance error, with no targets reached (0/256). (Right) Path length to target reveals that DARC maintains efficient trajectories, while PPO exhibits the longest paths.} %reflecting suboptimal trajectory planning.
    \label{fig:t-test}
\end{figure}

Since CESN is not provided with additional LfD data for the new target points, its performance on novel targets relies solely on its intrinsic extrapolation capability, which is limited to small extrapolations (\cite{cesn}). As a result, in this experiment, in contrast to reaching without an obstacle, CESN was unable to reach any of the extrapolated points (0/256). Curiously, its endpoint error was comparable to that of the first experiment, possibly, due to the fact in this experiment the robot state feedback was used as an input to the reservoir network. The DARC model,  by learning to modify context input to CESN, is able to reach successfully to all points (256/256).
% \erhan{In contrast to first experiment, PPO is allowed to generate torque output at each time-step}\erhcom{is it correct?}. 
Since PPO is directly trained on the novel targets, better performance than that of CESN might be expected.
% when it can issue torque commands at each time-step. 
Indeed, PPO achieves 171 successful episodes out of 256. However, under the same training conditions as the RL module in DARC, it exhibits a higher path-length error. DARC model seems to combine the strengths of LfD and RL and give the best performances (Figure~ \ref{fig:t-test}). 
% \erhcom{You have to explain how the path-length calculated and averaged for failed trials. Probably PPO and CESN would have worse path length if it were allowed to run 100 steps}
% Although the path length value for the CESN is lower than Darc, ...

\begin{figure}[ht]
    \centering
    \includegraphics[width=0.85\textwidth]{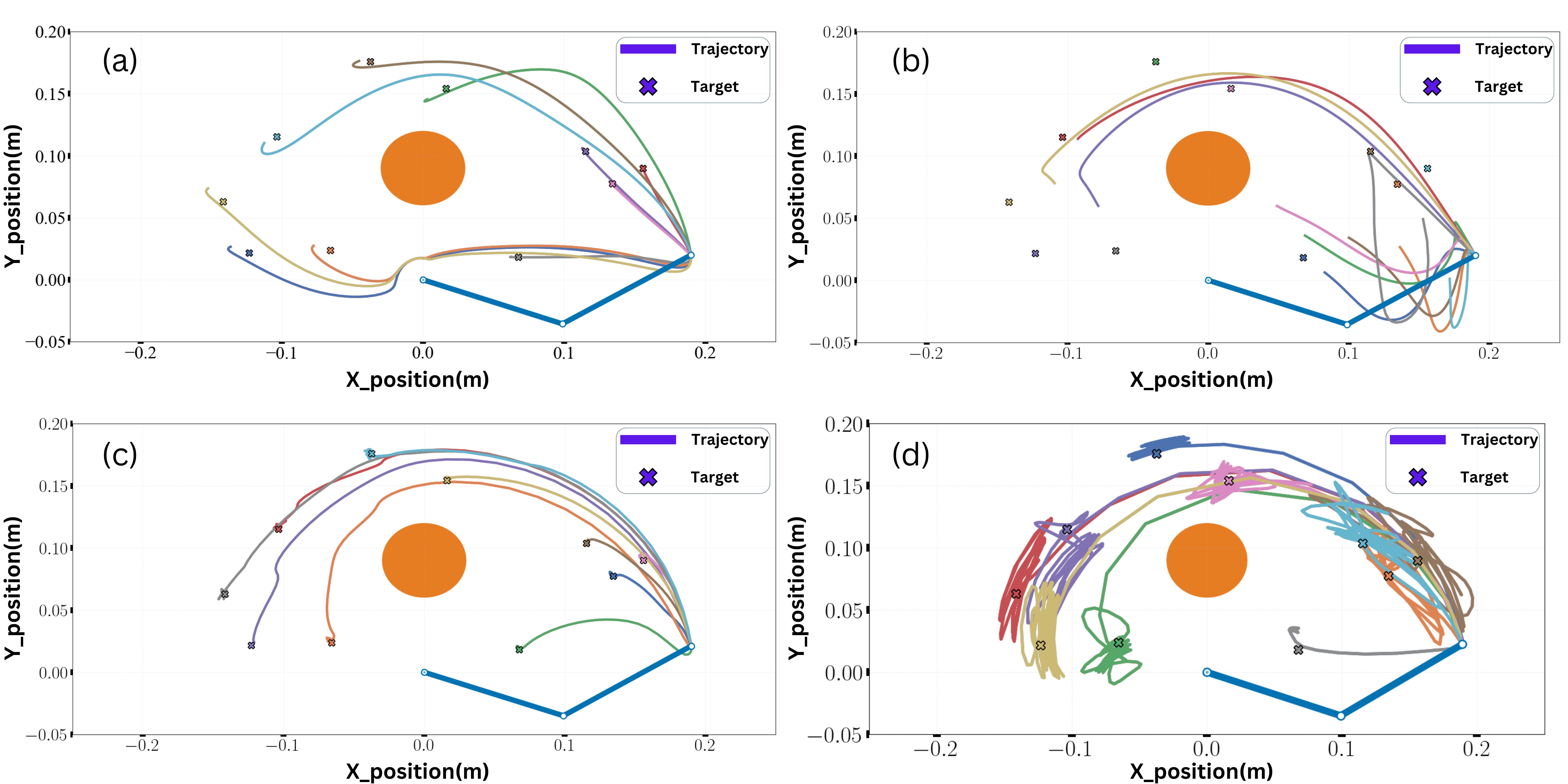}
    \caption{Comparison of 10 test trajectories for the Reaching with Obstacle Avoidance task. (a) Ground truth from PD controller. (b) CESN model. (c) DARC model. (d) PPO model. "X" marks target positions, and the orange circle represents the obstacle. Colored lines indicate each model's trajectory toward the targets while avoiding the obstacle.}
    \label{fig:med_trajs}
\end{figure}

Figure~\ref{fig:med_trajs} illustrates sample end-effector trajectories as each model attempts to reach various target points in the presence of an obstacle, shown in orange.
The CESN model, depicted in Figure~\ref{fig:med_trajs}b, struggles considerably as it relies only on its intrinsic extrapolation capacity.
% \sout{with obstacle avoidance and target accuracy in this task. Its trajectories are not only erratic but also frequently diverge from the intended path.}
The DARC model, on the other hand, shown in Figure~\ref{fig:med_trajs}c, achieves high accuracy and  obstacle avoidance by generating the context input during the episode at every 5 steps. 
% \erhcom{Don't say this now! It is your experiment plan to run at this frequency. So you must say it at the beginning: e.g. tell that "we aim to test DARC with RL running at 20Hz.." }. 
The standalone PPO model, presented in Figure~\ref{fig:med_trajs}d, shows partial success in reaching the target points. Although PPO demonstrates better performance than CESN, it struggles to maintain smooth trajectories. 
% \erhcom{what is the control frequency of PPO, how long it was trained etc. You need to report the details so that the results have a value. Think like this: The readers does not know whether you (on purpose or by mistake) run PPO to an insufficient level, or the network size were not suitable, etc. It is your duty to make readers understand that you allowed PPO to do its best; or it performed under your fair allotment (e.g., each model was given maximum of 10 hours of training time, or all models were trained until delta-loss was less than $1-e10$.)}

\subsubsection{\textbf{Novel Task of circular path tracking}}
This experiment aims to evaluate the transfer learning capability of our proposed model DARC. First, the reservoir is trained on demonstration data to reach static targets, same as the reaching task with obstacle avoidance.
% \erhcom{isn't exactly the same? Be precise, if different tell what is different}. 
Then, the RL module of DARC is trained to track a moving target along a specified circular path, while continuing to avoid obstacles accurately.
Then the robot is asked to follow a circular path centered at  [0,0.11] meters, with a radius of 7 cm, over a duration of 300 steps. In the case of CESN, the moving target is provided as context at each step, allowing for some tracking due on its extrapolation capability. In constrast both the PPO and DARC models, are allowed to learn and generate actions at every 5 steps. %, enabling them to periodically adjust their movements and stay aligned with the moving target along the path.

\begin{figure}[ht]
    \centering
    \includegraphics[width=\textwidth]{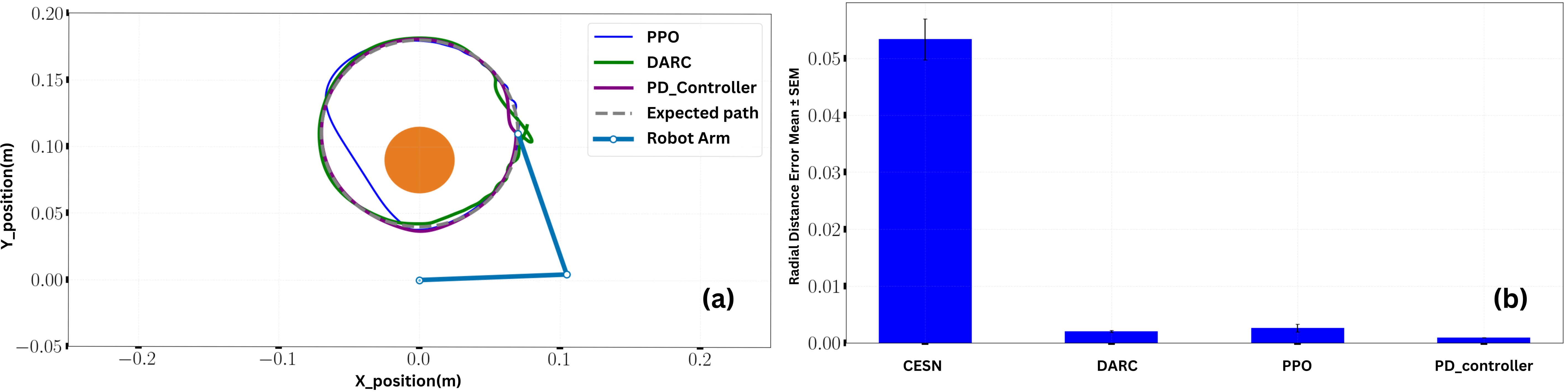}
    \caption{(a)  Example result of trajectory following task.  (b) Average radial distance error (mean ± SEM) for the trajectory following task across three training sessions for each model.}
    \label{fig:task_3}
\end{figure}

Figure~\ref{fig:task_3}b provides a quantitative comparison of the models in terms of radial distance error over three trials.
% \erhcom{how many?}. 
As expected, the CESN model shows the highest tracking error, with erratic and inconsistent movements. %, which aligns with its limitations in adapting to dynamic targets. I
In contrast, both DARC and the PPO controller achieve much better tracking (see  Figure~\ref{fig:task_3}a), achieving low radial distance errors, with a slight advantage for DARC. Notably, DARC’s performance is nearly on par with the baseline PD controller. %PPO’s error, while lower than CESN’s, is higher than that of DARC and the PD controller.

% \erhcom{Again MAIN result first} 
Figure~\ref{fig:task_3}.a shows an example trajectory for the circular path-following task in the presence of an obstacle (orange). The expected trajectory (dashed line) forms a smooth circle around the obstacle, and each model's performance in following this path is depicted. The PD controller (purple) serves as a benchmark, achieving near-perfect tracking by closely adhering to the expected path. The DARC model (green) also demonstrates effective tracking, maintaining a smooth and consistent trajectory that closely aligns with the expected path. The PPO model (blue), however, shows larger deviations from the target path, particularly during narrow area between obstacle and the base point.

% the curved segments \erhcom{everywhere is curved! Use better description}.
 
\section{Conclusion}
In this paper, we introduced a new adaptive movement primitive approach, called DARC, which integrates Learning from Demonstration (LfD) with Reinforcement Learning (RL) to enhance the generalization capability, computational efficiency, and scalability of LfD. DARC has been shown to outperform standalone LfD and pure RL in solving unseen tasks, demonstrating superior adaptation and computational efficiency. By learning to generate dynamic context for new situations, DARC enables successful generalization. Additionally, we leveraged reservoir dynamics modulation within DARC to perform transfer learning, allowing it to solve novel tasks with little additional training.  
This work provides a foundation for further advancements. For instance, although the reservoir predicts the fingertip position, this prediction is not currently utilized; integrating an additional controller to backup the torque output of DARC should improve  performance.  Additionally, refining the reward structure by incorporating terms related to the reservoir dynamics could further optimize performance. Finally, while this study was conducted in a simulation environment, future work should focus on validating the approach through experiments on real robotic systems.

\section{ACKNOWLEDGMENTS}
This work was supported by the project JPNP16007 commissioned by the New Energy and Industrial Technology Development Organization (NEDO) and by the Japan Society for the Promotion of Science KAKENHI Grant Number JP23K24926.

\bibliographystyle{unsrtnat}
\bibliography{ModulatingReservoirDynamics_arxiv}

 \newpage
\appendix

\section{Feedback Augmentation details}
\label{sec:A}
The feedback signal, \( f_t \), is augmented with noise such that \( \hat{f}_t = f_t + \epsilon \), where \( \epsilon \) is a noise vector. The noise \( \epsilon \) is drawn from a uniform distribution, with its lower and upper bounds adjusted based on the variability in the demonstration dataset, as shown in Table 2. The feedback was augmented with this noise for 10 times.

\begin{table}[ht]
\centering
\begin{tabular}{l c c}
\toprule
\textbf{feedback} & \textbf{Range} & \textbf{Noise Range (±5\%)} \\ 
\midrule
 ${q}_1$ & [-1.47, 1.73] (rad) & ±0.16 (rad) \\ 
 ${q}_2$ & [0.85, 3.0] (rad) & ±0.11 (rad) \\ 
 $\dot{q}_1$ & [-3.95, 9.11] (rad/s) & ±0.65 (rad/s) \\ 
 $\dot{q}_2$ & [-1.38, 6.09] (rad/s) & ±0.37 (rad/s) \\ 
\bottomrule
\end{tabular}
\caption{ feedback data augmentation at a 5\% noise level}
\end{table}

\end{document}